\pdfoutput=1

\documentclass[11pt]{article}

\usepackage[final]{acl}

\usepackage{graphicx}
\usepackage{times}
\usepackage{latexsym}
\usepackage{booktabs}
\usepackage{amsmath}
\usepackage{multirow}
\usepackage{pgfplots}
\usepackage{algorithm}
\usepackage{algorithmic}

\usepackage{pifont}
\usepackage[T1]{fontenc}

\usepackage[utf8]{inputenc}

\usepackage{microtype}

\usepackage{inconsolata}

\title{GKT: A Novel Guidance-Based Knowledge Transfer Framework For Efficient Cloud-edge Collaboration LLM Deployment}

\author{Yao Yao$^{1,2,3}$, Zuchao Li$^{4,*}$ and Hai Zhao$^{1,2,3,}$\thanks{$\ $  Corresponding author. This research was supported by the Joint Research Project of
Yangtze River Delta Science and Technology Innovation Community (No.
2022CSJGG1400), the National Natural Science Foundation of China (No. 62306216), the Natural Science Foundation of Hubei Province of China (No. 2023AFB816), the Fundamental Research Funds for the Central Universities (No. 2042023kf0133).} \\
$^{1}$Department of Computer Science and Engineering, Shanghai Jiao Tong University\\
$^{2}$Shanghai Key Laboratory of Trusted Data Circulation and Governance in Web3\\
$^{3}$Key Laboratory of Shanghai Education Commission for Intelligent Interaction\\
and Cognitive Engineering, Shanghai Jiao Tong University\\
$^{4}$National Engineering Research Center for Multimedia Software, \\
School of Computer Science, Wuhan University, Wuhan, 430072, P. R. China \\
{\tt yaoyao27@sjtu.edu.cn, zcli-charlie@whu.edu.cn,}\\
{\tt zhaohai@cs.sjtu.edu.cn}\\
}
\vspace{10pt}  
\begin{document}
\maketitle
\vspace{20pt}  
\begin{abstract}

The burgeoning size of Large Language Models (LLMs) has led to enhanced capabilities in generating responses, albeit at the expense of increased inference times and elevated resource demands. Existing methods of acceleration, predominantly hinged on knowledge distillation, generally necessitate fine-tuning of considerably large models,  posing a challenge for average users. Furthermore, present techniques for expediting inference and reducing costs operate independently. To address these issues, we introduce a novel and intuitive \textbf{G}uidance-based \textbf{K}nowledge \textbf{T}ransfer (GKT) framework. This approach leverages a larger LLM as a ``teacher'' to create guidance prompts, paired with a smaller ``student'' model to finalize responses. Remarkably, GKT requires no fine-tuning and doesn't necessitate the teacher and student models to have the same vocabulary, allowing for extensive batch generation to accelerate the process while ensuring user customization. GKT can be seamlessly integrated into cloud-edge collaboration architectures, and is versatile enough for plug-and-play application across various models.
When utilizing ChatGPT as teacher model and Llama2-70B as the student model, we can achieve 95.00\% of ChatGPT's performance at 52\% of the cost. The results highlight substantial enhancements in accuracy and processing speed on the GSM8K and CSQA datasets, surpassing the performance of using either the student or teacher models in isolation. Our code is publicly available at \href{https://github.com/Zoeyyao27/GKT}{https://github.com/Zoeyyao27/GKT}  

\end{abstract}

\section{Introduction}

\begin{figure}
    \centering
    \includegraphics[width=1\linewidth]{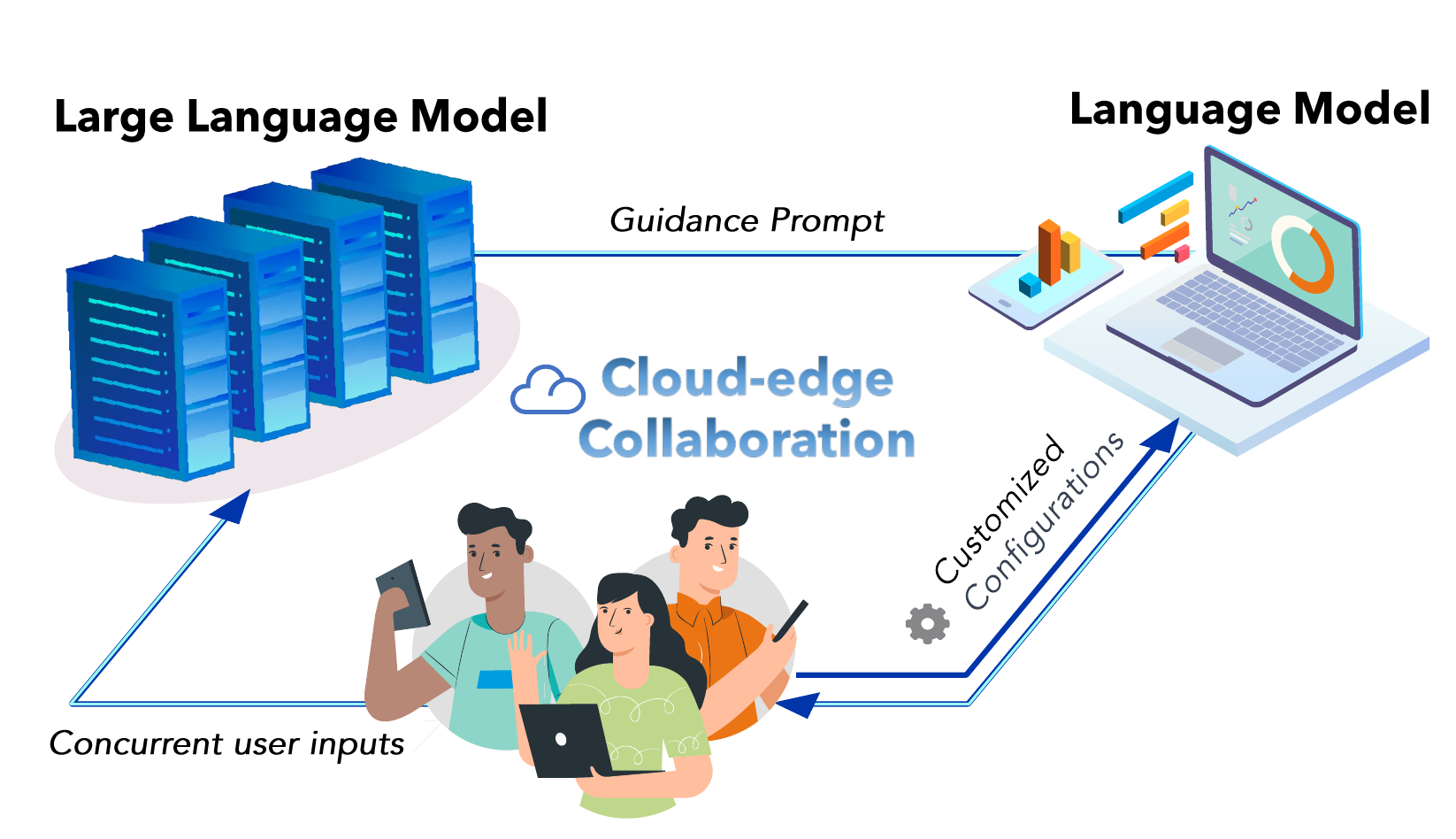}
    \caption{Cloud-edge collaboration: The GKT framework facilitates cloud-edge collaboration by deploying the larger teacher model on remote cloud servers and the smaller student model on lightweight mobile devices. GKT allows for brief guidance prompts to be easily transmitted to mobile devices, significantly reducing data transmission costs. In cloud-edge collaboration, users can also perform simple personalized generation settings on their mobile devices.}
    \label{fig:cloud}
\end{figure}

The swift advancement of large language models (LLMs) has dramatically pushed the frontiers of AI technology. LLMs, with their vast number of parameters, are exceptionally adept at comprehending human intentions, offering high-quality reasoning, and responses \cite{DBLP:conf/nips/Wei0SBIXCLZ22,DBLP:journals/corr/abs-2311-11797,DBLP:journals/corr/abs-2305-16582,DBLP:conf/nips/KojimaGRMI22}. However, the immense size of these models is a double-edged sword. While it improves model performance, it also leads to slower inference times and higher computational costs. As the demand for LLM usage increases, relying solely on LLMs for auto-regressive inference actually demands an overwhelming amount of computational resources and time, posing a substantial deployment challenges for cloud services and resource-constrained devices.

Consequently, many recent studies \cite{DBLP:conf/icml/LeviathanKM23,DBLP:journals/corr/abs-2307-15337,jiang2023lion} have taken steps to improve the inference efficiency of LLMs. One prevalent and widely adopted approach is knowledge distillation \cite{DBLP:journals/corr/HintonVD15, DBLP:conf/cvpr/YimJBK17, DBLP:journals/corr/abs-2310-16944,jiang2023lion,vicuna2023,li-etal-2023-symbolic}. 
The majority of knowledge distillation frameworks utilize large language models as ``teacher'' models to generate training samples. These samples are then used to train more compact ``student'' language models, effectively teaching them to mimic the performance and capabilities of teacher model. 
However, this process still demands a carefully crafted data generation mechanism and the subsequent training of the student model. Despite being smaller in size, to achieve satisfactory results in general tasks, many student models still maintain a considerable parameter size, often around 7B \cite{DBLP:journals/corr/abs-2310-16944} and 13B \cite{jiang2023lion,vicuna2023}. This is in line with the observations by \citet{DBLP:journals/tmlr/WeiTBRZBYBZMCHVLDF22}, who noted that the emergent abilities of LLMs for language understanding typically become evident when the model size exceeds 10 billion parameters.

\begin{table}[h]
\centering
\scalebox{0.8}{
\begin{tabular}{lccc}
\hline
Feature                           & KD & SD & GKT \\ \hline
Accelerates Inference             & \ding{51}                      & \ding{51}                    & \ding{51}   \\
Preserves Model Architecture      & \ding{51}                      & \ding{51}                    & \ding{51}   \\
Eliminates Additional Fine-tuning & \ding{55}                      & \ding{51}                    & \ding{51}   \\
Allows Custom Settings            & \ding{55}                      & \ding{55}                    & \ding{51}   \\
Enables Cloud-Edge Collaboration  & \ding{55}                      & \ding{55}                    & \ding{51}   \\
Allows Different vocabulary       & \ding{55}                      & \ding{55}                    & \ding{51}   \\\hline
\end{tabular}
}
\caption{Comparative analysis of Knowledge Distillation (KD), Speculative Decoding (SD), and Guidance-based Knowledge Transfer (GKT).}
\label{tab:comparison}
\end{table}

To solve this limitation, another research line focuses on speculative decoding \cite{DBLP:conf/icml/LeviathanKM23,DBLP:journals/corr/abs-2311-08252,pmlr-v202-leviathan23a}, which involves using a more efficient, smaller model to generate token predictions, which the larger target model then evaluates. If a token prediction is accepted, it's used; if not, it's discarded, and the target model generates a new token. The method is shown to significantly speed up inference without needing changes to the model's architecture or training procedures.
However, users often have diverse generation requirements, such as the desire to adjust generation parameters like temperature and top\_p. While speculative decoding can ensure consistency with the output of the final large model, employing batch generation to save time during high concurrent access can impede the ability to meet user-specific generation settings. 

\begin{figure*}[h]
    \centering
    \includegraphics[width=0.9\linewidth]{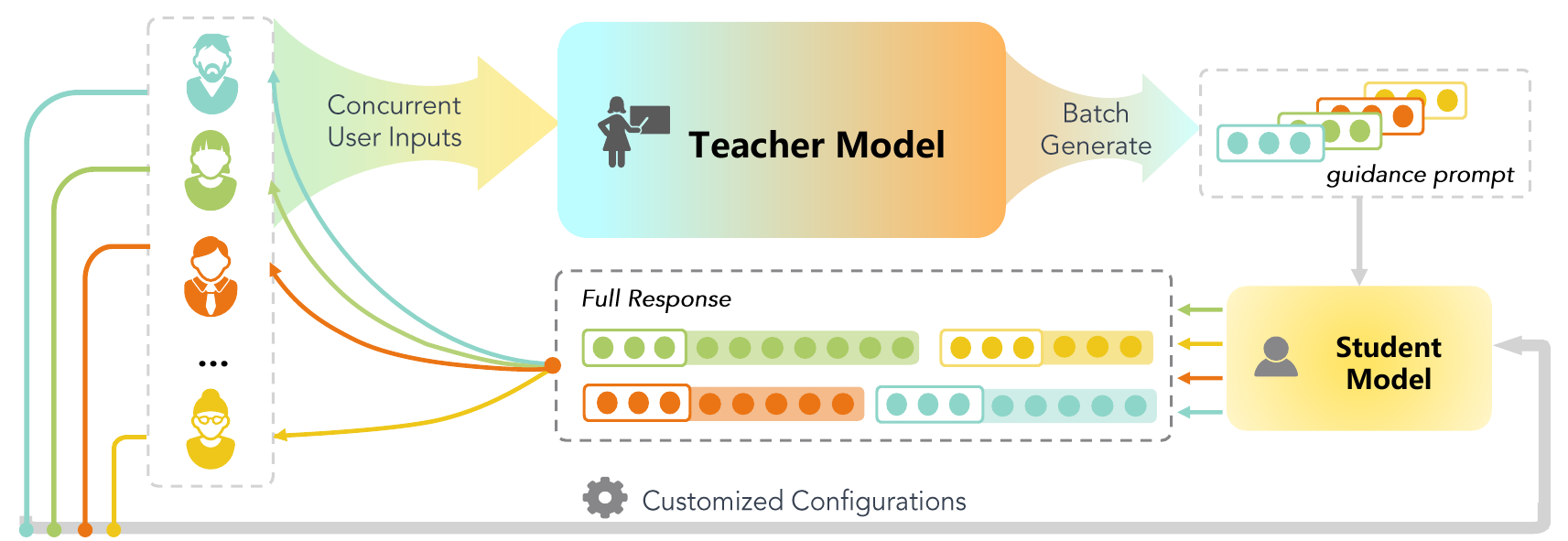}
    \caption{System overview. Our framework consists of two steps: guidance generation and response completion. In guidance generation, teacher model generates guidance prompts using batch generation to process concurrent user inputs. In response completion, student model receives guidance prompt and complete the response. Student model generates output with a batch size of 1 which allows customize generation settings by the user}
    \label{fig:overview}
\end{figure*}

Considering the limitations of these methods, we were inspired by the common human experience of ``Getting started is the hardest part.'' and sociological studies \cite{goldberg2014use} showing that effective prompts provided by teachers in classrooms can significantly improve student performance in exams. Thus, we propose a novel knowledge transfer framework: \textbf{G}uidance-based \textbf{K}nowledge \textbf{T}ransfer (GKT). Our framework involves two steps: firstly, using an LLM as a teacher model to generate guidance prompts from concurrent user inputs through batch generation. Secondly, a smaller LM acts as the student model, which simply completes the answers based on the guidance prompts, allowing for user-customized generation settings. 

GKT reduces the burden of LLM inference, thereby speeding up response generation. Unlike knowledge distillation, our framework does not require generating dataset from teacher model and fine-tuning the student model. It also doesn't necessitate the teacher and student models to have the same vocabulary, allowing for extensive batch generation to accelerate the process while ensuring user customization. GKT can also be seamlessly integrated into cloud-edge collaboration architectures as shown in Figure \ref{fig:cloud}. GKT deploys the larger teacher model on remote cloud servers and the smaller student model on lightweight mobile devices, such as smartphones. This setup allows for brief guidance prompts to be easily transmitted to mobile devices, significantly reducing data transmission costs. In cloud-edge collaboration, users can also perform simple personalized generation settings on their mobile devices. The table \ref{tab:comparison} succinctly delineates the pros and cons between the Guidance-based Knowledge Transfer (GKT) framework, Knowledge Distillation (KD) and Speculative Decoding (SD), providing a clear comparative perspective.

Finally, our study follows the philosophy that ``No such thing as bad student. Only bad teacher.'' We conducted various experiments to explore the optimal guidance strategies for the teacher model, seeking answers to questions such as how much and what type of help a teacher should provide to maximize the student model's accuracy while minimizing the teacher model's inference time. The results demonstrate that GKT not only achieves a significant accuracy improvement of 14.18\% on the GSM8K dataset, but also enhances speed by 10.72 $\times$. Furthermore, on the CSQA dataset, it records a noteworthy accuracy increase of 14.00\% along with a 7.73 $\times$ acceleration in inference speed.



\begin{table*}[!th]
\centering
\scalebox{0.7}{
\begin{tabular}{lllccccc}
\toprule
Dataset &Model                              & Output Length  &ACC$_{\text{teacher}}(\%)$                  & \textbf{ACC(\%)} & $\Delta(ACC)$ & Time(s)  & Speed Up      \\ \midrule
\multirow{16}{*}{GSM8K}& \multicolumn{6}{l}{\textit{Single Model}}                                                                                             \\
&Llama2-7B                          & 200 & -                           & 13.87   & -                          & 6945.70  & 1.31$\times$          \\
&                                   & 300 & -                           & 14.40   & -                          & 10304.18 & 1.38$\times$/13.98$\times$          \\
&Llama2-13B                         & 200 & -                           & 21.23   & 7.36                       & 9066.17  & -             \\
&                                   & 300 & -                           & 23.65   & 9.25                       & 14215.12 & 10.13$\times$             \\
&Llama2-70B                         & 300 & -                           & 56.63   & 42.23 / 32.98                       & 144018.55  & -             \\
&Bloom-3B                           & 300 & -                           & 2.35    & -                          & 4376.05  & 1.00$\times$          \\
&Bloom-7B                           & 300 & -                           & 4.40    & 2.05                       & 4392.74  & -             \\ \cline{2-8} 
&\multicolumn{6}{l}{\textit{GKT Framework}}                                                                                          \\
&Llama2-13B$\xrightarrow{}$Llama2-7B & 30$\xrightarrow{}$200           &\multirow{1}{*}{3.33} & 17.66   & 3.79                       & 7762.62  & 1.17$\times$          \\
&                                   & 30$\xrightarrow{}$300           & 3.33& 17.82   & 3.42                       & 10793.34 & 1.32$\times$          \\
&                                   & 40$\xrightarrow{}$300            &\multirow{1}{*}{4.62} & 19.18   & 4.78                       & 10871.01 & 1.31$\times$          \\
&                                   & 40(concise)$\xrightarrow{}$300 & 4.62& 19.26   & 4.86                       & 10707.43 & 1.33$\times$ \\
&Llama2-70B$\xrightarrow{}$Llama2-7B  & 40$\xrightarrow{}$300          & 7.81& 28.58    & \textbf{14.18}                      & 13440.71  & \textbf{10.72 $\times$ }        \\
&Llama2-70B$\xrightarrow{}$Llama2-13B  & 40$\xrightarrow{}$300          & 7.81& 35.41    & 11.76                      &  16250.34  & 8.86$\times$         \\
&Bloom-7B$\xrightarrow{}$Bloom-3B    & 40$\xrightarrow{}$300           & 2.05 & 2.58    & 0.23                       & 5198.60  & 0.84$\times$          \\
&Llama2-7B$\xrightarrow{}$Bloom-3B   & 40$\xrightarrow{}$300           &3.71 & 6.82    & 4.47                       & 4154.40  & 1.06$\times$          \\
&Llama2-13B$\xrightarrow{}$Bloom-3B  & 40$\xrightarrow{}$300          &\multirow{1}{*}{4.62} & 7.73    & 5.38                       & 4186.22  & 3.40$\times$          \\
&Llama2-13B$\xrightarrow{}$Bloom-7B  & 40$\xrightarrow{}$300          & 4.62& 10.31    & 5.91                       & 3275.90  & 3.82$\times$         \\
\midrule
\multirow{19}{*}{CSQA}&\textit{Single Model}              &                       &         &                            &          &          \\
&Llama2-7B                          & 100 & -                   & 60.69   & -                          & 3239.04  & 1.31 $\times$ / 13.36$\times$   \\
&                                   & 300 & -                  & 60.61   & -                          & 9321.51  & 1.30$\times$    \\
&Llama2-13B                         & 100 & -                  & 71.17   & 10.48                      & 4235.46  & 10.22$\times$        \\
&                                   & 300 & -                  & 71.09   & 10.48                      & 12128.50 & -        \\
&Llama2-70B                         & 100 & -                           & 76.58   & 15.89 / 5.41                       & 43285.73  & -             \\
&Bloom-3B                           & 100 & -                  & 20.88   & -                           & 2157.97  & 1.05$\times$    \\
&Bloom-7B                           & 100 & -                  & 21.79   & -                           & 2269.35  & -        \\ \cline{2-8} 
&\textit{GKT Framework}           &                       &         &                            &          &          \\
&Llama2-13B$\xrightarrow{}$Llama2-7B & 10$\xrightarrow{}$100  &0.00& 61.02   & 0.33                       & 3472.33  & 1.22$\times$    \\
&                                   & 20$\xrightarrow{}$100  &0.00& 64.70   & 4.01                       & 3656.64  & 1.16$\times$    \\
&                                   & 30$\xrightarrow{}$100  &18.76& 69.86   & 9.17                       & 3579.21  & 1.18$\times$    \\
&                                   & 30 $\xrightarrow{}$ 300  &18.76& 69.86   & 9.25                       & 10171.32 & 1.19$\times$    \\
&                                   & 40 $\xrightarrow{}$300 &62.74& 71.01   & 10.40                      & 10068.97 & 1.20$\times$    \\
&                                   & 50 $\xrightarrow{}$300  &70.43& 71.09   & 10.48                      & 10361.78 & 1.17$\times$    \\
&Llama2-70B$\xrightarrow{}$Llama2-7B  & 30$\xrightarrow{}$100          & 18.84& 74.69    & 14.00                      & 5600.38  & \textbf{7.73$\times$}         \\
&Llama2-70B$\xrightarrow{}$Llama2-13B  & 30$\xrightarrow{}$100          & 18.84& 76.16    & 4.99                      & 6598.43  & 6.56$\times$         \\
&Bloom-7B$\xrightarrow{}$Bloom-3B    & 20$\xrightarrow{}$100  &0.00& 20.96   & 0.08                       & 2395.36  & 0.95$\times$    \\
&                                   & 30$\xrightarrow{}$100  &11.06& 20.63   & -0.25                      & 2250.04  & 1.01$\times$    \\
&Llama2-7B$\xrightarrow{}$Bloom-3B   & 30$\xrightarrow{}$100   &21.86& 40.05   & 19.17                      & 2134.58  & 1.52$\times$    \\
&Llama2-13B$\xrightarrow{}$Bloom-3B  & 30$\xrightarrow{}$100  &18.76& 41.20   & \textbf{20.32}                      & 2210.29  & 1.92 $\times$    \\
&Llama2-13B$\xrightarrow{}$Bloom-7B  & 30$\xrightarrow{}$100  &18.76& 39.80   & 18.01                      & 2405.20  & 1.76$\times$    \\ \hline
\end{tabular}
}
\caption{Results for GSM8K and CSQA. ``$\xrightarrow{}$'' signifies the transition from the teacher model to the student model, with settings on the left of the arrow (Model, Output Length) pertaining to the teacher model, and those on the right corresponding to the student model. ``ACC$_{\text{teacher}}(\%)$'' and ``ACC(\%)'' denotes the mean accuracy(\%) achieved by the teacher model alone and the overall framework, respectively. ``$\Delta$(ACC)'' denotes the improvement in accuracy (\%) achieved by the GKT framework compared to using only the student model ($\Delta$(ACC) for Llama2-70B shows two numbers separated by ``/'', The number on the left (right) of the ``/'' is the change in accuracy when using Llama2-7B (Llama2-13B) as the student model). ``Speed Up'' indicates the acceleration factor of the GKT framework relative to using only the teacher model. (``Speed Up'' for Llama2-7B shows two numbers separated by ``/'', The number on the left (right) of the ``/'' is the change in accuracy when using Llama2-13B (Llama2-70B) as the student model). ``concise'' denotes that we use concise guidance generation method and the detailed analysis can be found in section \ref{sec:teach_method} }
\label{tab:overall result}
\end{table*}

\section{Method}
\label{sec:method}

The schematic representation of our Guidance-Based Knowledge Transfer (GKT) framework is depicted in Figure \ref{fig:overview}. GKT encompasses a two-step process: guidance generation and response completion. During the guidance generation phase, a large language model serves as the ``teacher'' model. This model processes concurrent user inputs and employs batch generation to craft guidance prompts. Subsequently, in the response completion stage, a smaller language model functions as the ``student'' model. This model offers flexibility in generation settings, allowing for user customization. The guidance prompt created in the first stage is then fed into this smaller model, facilitating the completion of the response with enhanced efficiency.
We will now elaborate on each stage:

\paragraph{Guidance Generation} 
Given concurrent user inputs \( Q = \{q_1, q_2, ..., q_n\} \), where \( q_i \) represents the input question from user \( i \), the teacher model \( M_t \) batch generates the guidance prompts:
$$ G = \{g_1, g_2, ..., g_n\} = \mathbf{F}(M_t(Q)) $$
Here, \( G \) represents the batch-generated guidance, \( g_i \) is the guidance for user \( i \), and \( \mathbf{F}(\cdot) \) denotes the projection operation that generates the guidance prompt from the generated text. In this paper, we explore different projection operations including: (1) Cut-off guidance generation (2) Concise guidance generation (3) Hint guidance generation. We will elaborate on these methods in Section \ref{sec:teach_method}

\paragraph{Response generation}
In response generation, we use a smaller language model $M_s$ as student model. For every user $i$, $M_s$ generates the final response $r_i$ by:
$$r_i=M_s^i(g_i)$$
Where $M_s^i$ stands for $M_s$ under the user $i$'s custom generation setting. Figure \ref{fig:GKT_example} in Appendix \ref{append:GKT_example} demonstrates a concrete example for GKT framework.

\section{Experimental Settings}
\paragraph{Datasets} In this paper we use two challenge but widely used dataset: GSM8K \cite{DBLP:journals/corr/abs-2110-14168} for arithmetic reasoning and CSQA \cite{talmor-etal-2019-commonsenseqa} for commonsense reasoning. GSM8K focus on arithmetic reasoning which is a collection of grade school math word problems, each requiring 2 to 8 steps to solve. The solutions mainly involve a series of basic arithmetic calculations to arrive at the final answer. CSQA focuses on commonsense question answering which includes multiple-choice questions that require commonsense knowledge for answering. The detailed dataset statistics can be found in Appendix \ref{append:dataset}. In all our experiments, we use the same prompt settings in Manual-CoT \cite{DBLP:conf/nips/Wei0SBIXCLZ22} otherwise stated. The full prompt can be found in Appendix \ref{append:prompt}

\paragraph{Hyperparameter}
All experiments were conducted on an NVIDIA A800 GPU. Detailed hyperparameter settings for the experiments are provided in Appendix \ref{append:param}. In our experiments, we tested various teacher models, including Flan-t5-xl \cite{DBLP:journals/corr/abs-2211-05100}, Bloom-7B\cite{DBLP:journals/corr/abs-2211-05100}, Llama2-70B\cite{DBLP:journals/corr/abs-2307-09288} and Llama2-13B. Correspondingly, the student models used were Flan-t5-large, Bloom-3B, Llama2-13B and Llama2-7B.

\section{Results and Exploration}
In this section, we delve into the empirical results of our comprehensive analysis using the GKT framework. The essence of this exploration lies in quantifying the effectiveness of GKT in enhancing the accuracy of student models while ensuring computational efficiency.
We first report the overall results for the GKT on the GSM8K and CSQA datasets. Then, we explore various dimensions of knowledge transfer including the optimal guidance generation methods, the intriguing dynamics between different types of teacher and student models, the influence that a teacher can exert on student and the optimal guidance length. 

\subsection{Overall Results}
The overall results for the Guidance-based Knowledge Transfer (GKT) framework on the GSM8K and CSQA datasets are presented in Table \ref{tab:overall result}. The results indicate that on the GSM8K dataset, we achieved a maximum accuracy improvement of 14.18 \% compared to simply using the student model, along with a 10.72$\times$ speed up relative to the teacher model. On the CSQA dataset, we observed a accuracy improvement of 14.00 \% compared to the student model, and a maximum 7.73$\times$ speed-up in comparison to the teacher model.

\subsection{How to facilitate student learning effectively?}
\label{sec:teach_method}
\begin{table*}[t]
\scalebox{0.7}{
\begin{tabular}{lllllll}
\toprule
Model                                               & Prompt                                    & Output   Length                & \textbf{ACC(\%)} & $\Delta(ACC)$ & Time(s)  & Speed Up \\ \hline
\multicolumn{7}{l}{\textit{Single Model}}                                                                                                                                                              \\
Llama2-7B                                           & -                                         & 300                            & 14.40   & -                          & 10304.18 & 1.38$\times$    \\
Llama2-13B                                          & -                                         & 300                            & 23.65   & 9.25                       & 14215.12 & -        \\\hline
\multicolumn{7}{l}{\textit{GKT Framework}}                                                                                                                                                           \\
\multirow{3}{*}{Llama2-13B$\xrightarrow{}$Llama2-7B} & -                                         & 40$\xrightarrow{}$300           & 19.18   & 4.78                       & 10871.01 & 1.31$\times$    \\
                                                    & "Provide the answer in a brief manner: "  & 40(concise)$\xrightarrow{}$300 & 19.26   & 4.86                       & 10707.43 & 1.33$\times$    \\
                                                    & "Provide a brief hint for the question: " & 40(hint)$\xrightarrow{}$300     & 19.11   & 4.71                       & 11190.12 & 1.27$\times$ 
                                                   \\ \hline
\end{tabular}
}

\caption{Results of different guidance generation methods on GSM8K. ``concise'' and ``hint'' denotes Concise Guidance Generation and Hint Guidance Generation respectively}
\label{tab:teach_method}
\end{table*}

To investigate how to generate better guidance to assist student models in answering questions, we experimented with three different guidance generation methods as described in Section \ref{sec:method}: (1) Cut-off Guidance Generation, (2) Concise Guidance Generation, and (3) Hint Guidance Generation.

\textbf{Cut-off Guidance Generation}: We employed the simplest method of cutting off, where the teacher model generates only a fixed number of the first $m$ tokens as guidance.

\textbf{Concise Guidance Generation}: In this approach, we added the prompt: ``Provide the answer in a brief manner:'' to guide the model to generate more concise guidance responses.

\textbf{Hint Guidance Generation}: Here, we introduced the prompt ``Provide a brief hint for the question:'' to encourage the teacher model not to give direct answers but to offer hints in a guiding manner, aiding the student model in generating responses.

The detailed results of these experiments on GSM8K can be seen in Table \ref{tab:teach_method}. To more intuitively understand the acceleration effect of each component in the GKT framework, we have created a trace diagram of GKT's performance on the GSM8K dataset, as illustrated in Figure \ref{fig:teach_method}. Intriguingly, we found that providing hints leads to poorer outcomes compared to directly giving the answer. We speculate that this may be due to the limited inferential and reasoning capabilities of the smaller models. \textbf{Instead of giving hints for them to infer, it might be more effective to provide direct answers.} We also observed that Concise Guidance Generation was relatively effective, as the brevity of the guidance reduces the inferential workload for the student model. By prompting the teacher model to produce shorter answers, this method not only improved the accuracy of the model's responses but also accelerated the inference speed.

\subsection{GKT for Cloud-edge Collaboration LLM Deployment}
\begin{figure*}[h]
    \centering
    \includegraphics[width=0.75\linewidth]{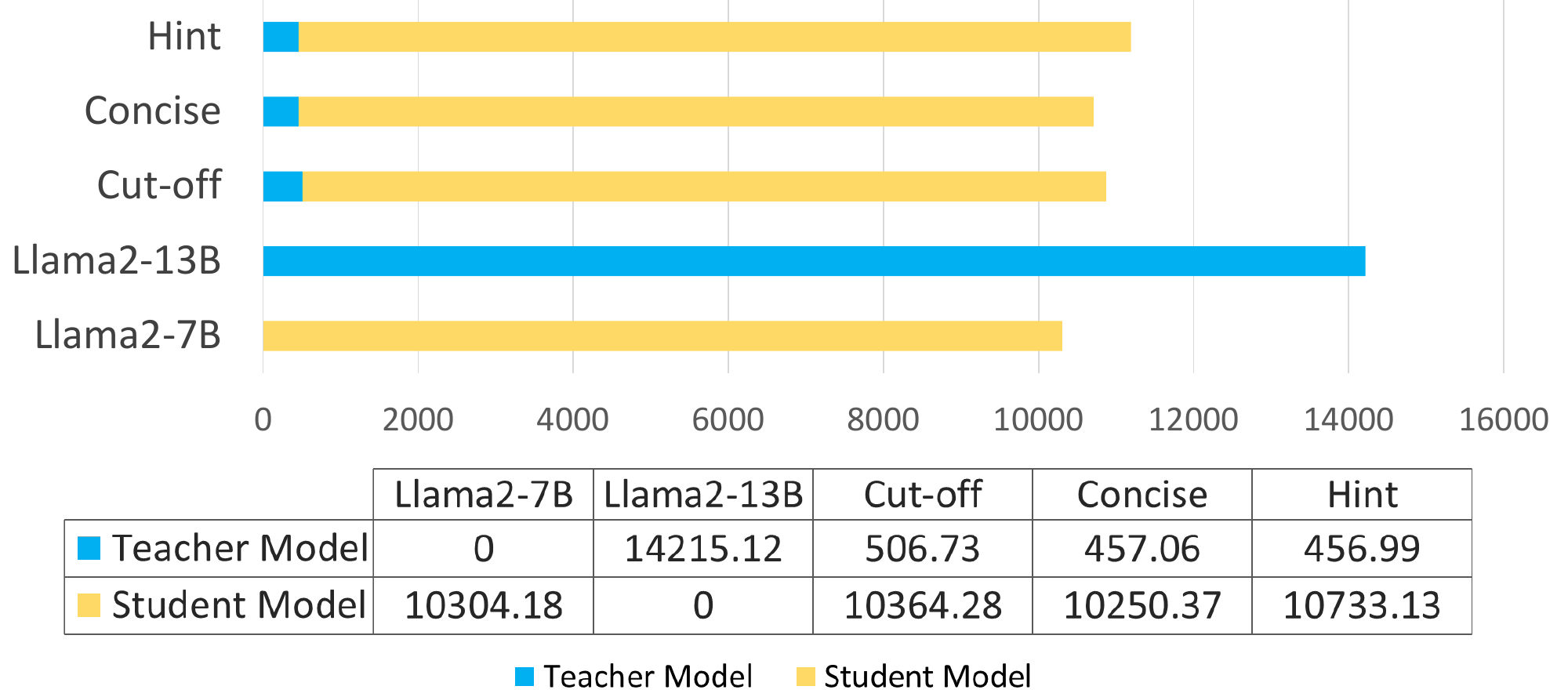}
    \caption{The trace diagram for GKT framework on GSM8K. The vertical axis represents the guidance generation method used (where ``Llama2-13B'' indicates the exclusive use of the Llama2-13B model for response generation, and ``Llama2-7B'' denotes the sole use of the Llama2-7B model). The horizontal axis represents the runtime (s) of the GKT across the entire dataset.}
    \label{fig:teach_method}
\end{figure*}

Figure \ref{fig:teach_method} demonstrates that utilizing only the Llama2-13B model for the GSM8K dataset, comprising a total of 1319 examples, results in an average response time of 10.78 seconds per example ($14215.12 \div 1319=10.78s$). This implies a single-user service capability within this timeframe. Conversely, the deployment of the GKT framework in a Cloud-Edge collaboration environment for LLMs markedly reduces the large model's response time to 0.38 seconds on average ($506.73 \div 1319=0.38s$), and the small model processes each example in 7.86 seconds ($10364.28 \div 1319=7.86s$), culminating in a total response time of 8.24 seconds. Therefore, theoretically, by employing batch processing, \textbf{the GKT framework can facilitate simultaneous service to 24 users within the 8.24-second window. }
This efficiency stems from the large model's capacity for batch processing in the cloud, which can concurrently serve multiple users (with a batch size of 24). At the same time, the small models are deployed on distinct edge devices, facilitating parallel operations and enabling personalized user experiences.
\textbf{In stark contrast, reliance solely on the Llama2-13B model limits service to a single user within the 10.78-second timeframe.} Hence, the GKT framework substantially augments the parallelism in user service provision in Cloud-Edge collaborative LLM deployments.

Additionally, compared to speculative decoding, GKT can control transmission losses in low-bandwidth environments by adjusting the length of the guidance prompt. We provide a comprehensive analysis of GKT and speculative decoding in Appendix \ref{append:vs}.

\subsection{How To Identify the Right Teacher for the Right Student ?}
To find the right student for the right teacher, we experimented with the decoder-only model Llama and the encoder-decoder model Flan-t5. Our experiments show that both types of models can achieve certain improvements through GKT. However, an intriguing phenomenon we observed was that replacing the teacher model of Flan-t5 with a larger Llama model led to an accuracy decrease, rather than an increase. We speculate that one possible reason could be that for encoder-decoder models, the encoder and decoder use different parameters. Using a large model's answers as input for the smaller model's encoder may disrupt the coherence of thought, leading to reduced inferential ability. In contrast, for decoder-only models, directly concatenating inputs and targets seems to aid in better inference. The significant structural differences between T5 and Llama models could result in inconsistent thinking patterns between models, implying that the teacher and student are not perfectly in sync, thereby diminishing the overall effectiveness.
\begin{table*}[h]
\centering
\scalebox{0.75}{
\begin{tabular}{llllll}
\toprule
Model                                  & Output   Length      & \textbf{ACC(\%)} & $\Delta(ACC)$ & Time(s)  & Speed Up \\ \midrule
\multicolumn{6}{l}{\textit{Single Model}}                                                                                  \\
Flan-t5-large (0.8B)                   & 300                  & 6.90    & -                          & 2343.06  & 1.09$\times$    \\
Flan-t5-xl (3B)                        & 300                  & 11.30   & 4.40                       & 2557.76  & -        \\
Llama2-7B                              & 300                  & 14.40   & -                          & 10304.18 & 1.38$\times$    \\
Llama2-13B                             & 300                  & 23.65   & 12.35                      & 14215.12 & -        \\ \midrule
\multicolumn{6}{l}{\textit{GKT Framework}}                                                                               \\
Flan-t5-xl$\xrightarrow{}$Flan-t5-large & 40$\xrightarrow{}$300 & 7.88    & 0.98                       & 2190.05  & 1.17$\times$    \\
Llama2-13B$\xrightarrow{}$Llama2-7B     & 40$\xrightarrow{}$300 & 19.18   & 4.78                       & 10871.01 & 1.31$\times$    \\
Llama2-13B$\xrightarrow{}$Flan-t5-large & 40$\xrightarrow{}$300 & 6.52    & -0.38                      & 2330.48  & 6.10$\times$    \\
Llama2-7B$\xrightarrow{}$Flan-t5-large  & 40$\xrightarrow{}$300 & 6.21    & -0.69                      & 2284.67  & 4.51$\times$    \\ \bottomrule
\end{tabular}
}
\caption{Decoder-only Model VS Encoder-Decoder Model. We experimented with the decoder-only model Llama and the encoder-decoder model Flan-t5. }
\label{tab:right_teacher_right_student}
\end{table*}

\subsection{How Can a Teacher Influence His Student?}
Based on the above findings, we conducted further experiments to explore the impact of a teacher's guidance on student models, specifically focusing on decoder-only models. In these experiments, we used Llama2 and Bloom, both decoder-only models. The overall results can be seen in Table \ref{tab:Teacher-influence}. 
Table \ref{tab:Teacher-influence} presents a detailed overview of the impact of teacher model choice on the performance and efficiency of student models, as tested on the GSM8K and CSQA datasets. The results highlight the differential effects on accuracy and processing speed, depending on the combination of teacher and student models used. The results also revealed that the Llama model possesses stronger inferential capabilities and stores more common-sense knowledge, performing better on both datasets. When we provided the same Bloom student model with the more experienced teacher model Llama, under similar-sized teacher model conditions, we observed that the Llama model, as a teacher, could enhance the student model's accuracy by 20 \%. In contrast, using Bloom-7B as the teacher resulted in a decrease in accuracy. This outcome underscores the critical influence of the teacher on the student.

Additionally, by comparing results on the GSM8K and CSQA datasets, we found that this approach of using larger models to guide smaller ones can better transfer the common-sense knowledge stored in the teacher model. However, it had a less pronounced effect on mathematical reasoning abilities in the GSM8K dataset. On the CSQA dataset, when employing a Llama2-7B teacher model, Bloom-3B model's accuracy is improved by nearly 20 \%, with a 1.52 times speed increase. These results convincingly demonstrate that our method can effectively transfer the knowledge and reasoning abilities stored in larger models to smaller ones.

\begin{table*}[h]
\centering
\scalebox{0.75}{
\begin{tabular}{llllll}
\toprule
Model                              & Output   Length      & \textbf{ACC(\%)} & $\Delta(ACC)$  & Time(s)  & Speed Up \\ \midrule
\multicolumn{6}{l}{\textit{GSM8K}}                                                                                     \\
Llama2-13B$\xrightarrow{}$Llama2-7B & 40$\xrightarrow{}$300 & 19.18   & 4.78                       & 10871.01 & 1.31$\times$    \\
Bloom-7B$\xrightarrow{}$Bloom-3B    & 40$\xrightarrow{}$300 & 2.58    & 0.23                       & 5198.60  & 0.84$\times$    \\
Llama2-7B$\xrightarrow{}$Bloom-3B   & 40$\xrightarrow{}$300 & 6.82    & 4.47                       & 4154.40  & 1.06$\times$    \\
Llama2-13B$\xrightarrow{}$Bloom-3B  & 40$\xrightarrow{}$300 & 7.73    & 5.38                       & 4186.22  & 3.40$\times$    \\
Llama2-13B$\xrightarrow{}$Bloom-7B  & 40$\xrightarrow{}$300 & 10.31   & 5.91                       & 3725.90  & 3.82$\times$    \\ \midrule
\multicolumn{6}{l}{\textit{CSQA}}                                                                                      \\
Llama2-13B$\xrightarrow{}$Llama2-7B & 30$\xrightarrow{}$100 & 69.86   & 9.17                       & 3579.21  & 1.18$\times$    \\
Bloom-7B$\xrightarrow{}$Bloom-3B    & 30$\xrightarrow{}$100 & 20.63   & -0.25                      & 2250.04  & 1.01$\times$    \\
Llama2-7B$\xrightarrow{}$Bloom-3B   & 30$\xrightarrow{}$100 & 40.05   & 19.17                      & 2134.58  & 1.52$\times$    \\
Llama2-13B$\xrightarrow{}$Bloom-3B  & 30$\xrightarrow{}$100 & 41.2    & 20.32                      & 2210.29  & 1.92$\times$    \\
Llama2-13B$\xrightarrow{}$Bloom-7B  & 30$\xrightarrow{}$100 & 39.8    & 18.01                      & 2405.2   & 1.76$\times$    \\ \bottomrule
\end{tabular}
}
\caption{Comparative analysis of teacher influence on student models using decoder-only models, Llama and Bloom.}
\label{tab:Teacher-influence}
\end{table*}

\begin{table}[!h]
\centering
\scalebox{0.75}{
\begin{tabular}{ccccc}
\toprule
                       & \multicolumn{2}{l}{Average Output Length} & \multirow{2}{*}{\textbf{ACC}} & \multirow{2}{*}{$\Delta(ACC)$} \\ \cline{2-3}
                       & ChatGPT                        & Ours         &                      &                        \\ \midrule
Full ChatGPT & \multirow{1}{*}{76.68}         &           - & 68.16                & -                  \\\midrule
\multirow{4}{*}{Llama2-7B}                       &                                -& 10 $\xrightarrow{}$ 300           & 19.41                & 0.23                   \\
                       &                                -& 20 $\xrightarrow{}$ 300         & 27.52                & 8.34                   \\
                       &                                -& 30 $\xrightarrow{}$ 300          & 32.83                & 13.65                  \\
                       &                                -& 40 $\xrightarrow{}$ 300          & 40.79                & 21.61                  \\ 
\multirow{1}{*}{Llama2-13B}                
                       &                                -& 40 $\xrightarrow{}$ 300          & 48.14                & 24.49                  \\ 
\multirow{1}{*}{Llama2-70B} 
                       &                                -& 40 $\xrightarrow{}$ 300          & 64.75                & 8.12                  \\ \bottomrule

\end{tabular}
}
\caption{Result on GSM8K when using ChatGPT as teacher model and Llama2-7B, 13B, 70B as student model. ``$\xrightarrow{}$'' signifies the output length transition from ChatGPT to the Llama2 model. ``$\Delta$(ACC)'' denotes the improvement in accuracy (\%) achieved by GKT framework compared to using only the corresponding Llama2 model. ``Full ChatGPT'' denotes the ChatGPT performance on GSM8K without GKT}
\label{tab:chatgpt-teacher}
\end{table}

To delve deeper into the extent of influence a teacher model can have on a student model, we utilized the superior-performing ChatGPT \cite{DBLP:journals/corr/abs-2303-08774} as the teacher model, with Llama2-7B, 13B, and 70B serving as the student model. The Table \ref{tab:chatgpt-teacher} clearly demonstrates that using a more powerful teacher model significantly enhances performance. When the number of guidance tokens is kept constant, employing ChatGPT as the teacher model results in a substantial improvement over using Llama2-13B as the teacher model. Specifically, on the GSM8K dataset, there's an increase of 21.53 \% compared to Llama2-13B teacher (from 19.26\% to 40.79\% ). We can also see from the table that stronger student model can preserve more teacher model abilities. From another perspective, this approach allows for more cost-effective outcomes. When utilizing ChatGPT’s API interface and employing the best performance Llama2-70B as the student model, we can achieve 95.00\% of ChatGPT's performance at 52\% of the cost, effectively showcasing the GKT framework's ``cheap and cheerful'' charm.


\begin{table*}[t]
\centering
\scalebox{0.75}{
\begin{tabular}{llccccc}
\toprule
Model                  & Output Length   & ACC$_{\text{teacher}}$(\%) & \textbf{ACC(\%)}   & $\Delta(ACC)$ & Time(s) & Speed up \\ \midrule
Llama2-7B              & 100             & -            & 27.17 & -          & 707.02  & 13.05$\times$    \\
Llama2-70B             & 100             & -            & 40.16 & 12.99      & 9224.76 & -        \\ \midrule
Llama2-70B$\xrightarrow{}$ Llama2-7B & 20$\xrightarrow{}$100 & 5.51         & 30.31 & 3.14       & 1070.02 & 8.62$\times$     \\ \bottomrule
\end{tabular}}
\caption{Results on AQuA-RAT dataset}
\label{tab:auqa}
\end{table*}

\subsection{Further Exploration}
\subsubsection{Results on additional datasets}
To better demonstrate the effectiveness of GKT across different datasets, we conducted supplementary experiments on the AQuA-RAT \cite{DBLP:conf/acl/LingYDB17} dataset. The results for the AQuA-RAT dataset are presented in the Table \ref{tab:auqa} which shows that GKT also exhibits superior capabilities on the AQuA-RAT dataset, powerfully affirming the effectiveness of the GKT framework and its robustness across various datasets.

\begin{table}[h]
\centering
\scalebox{0.75}{
\begin{tabular}{lllll}
\toprule
Model                  & Rouge  & $\Delta$(Rouge) & Time(s) \\ \midrule
Llama2-7B              & 0.1475 & -           & 10304.18     \\
Llama2-13B             & 0.1631 & 0.0156      & 14215.12 \\ \midrule
Llama2-13B$\xrightarrow{}$Llama2-7B& 0.1578 & 0.0103      & 10871.01 \\ \bottomrule
\end{tabular}}
\caption{Rouge scores on GSM8K rationale dataset. We set the output length for the teacher model to 40 tokens and for the student model to 300 tokens.}
\label{tab:rouge_result}
\end{table}

To validate the performance of the GKT framework in more open-ended generative tasks, we also calculated and compared the Rouge scores of the rationales generated by the GKT model on the GSM8K dataset. The GSM8K dataset contains two types of labels. The first is the rationale for each question, which is a text string providing a solution to the question. The second label is the numeric format answer to the question. Examples of the GSM8K dataset can be found in Figure \ref{fig:prompt_GSM8K}, where the blue part is the rationale, and the pink part is the answer. In our evaluation, we expanded beyond just assessing the final numeric format answers (Accuracy) by also evaluating the rationale provided for answering the question (Rouge). This approach helps in assessing how well the GKT framework adapts to tasks that require extensive content generation. The results in terms of Rouge scores are presented in the Table \ref{tab:rouge_result}.

\subsubsection{How much guidance should a teacher offer to students ?}
\label{append:guidance_len}
\begin{figure}[!h] 
\centering 

\begin{tikzpicture} 

\begin{axis}[
    width=6cm, 
    height=4cm, 
    xlabel=Guidance Length, 
    ylabel= Accuracy(\%), 
    tick align=inside, 
    legend columns=-1,
    legend style={at={(0.5,1.1)},anchor=south}, 
    minor tick num=1,
    grid=both,
    grid style=dashed
    ]

\addplot[color=cyan,mark=*,mark size=1.5pt,] coordinates{(10,14.78)(20,17.82)(30,17.82)(40,19.18)};
\addlegendentry{$\textrm{GSM8K}$}
\addplot[color=magenta,mark=*,mark size=1.5pt,] coordinates{(10,61.02)(20,64.70)(30,69.86)(40,71.01)};
\addlegendentry{$\textrm{CSQA}$}

\end{axis}
\end{tikzpicture}
\caption{Performance of different length ranges. We use Llama2-13B as teacher model and Llama2-7B as student model. }
\label{fig:length_range}
\end{figure}
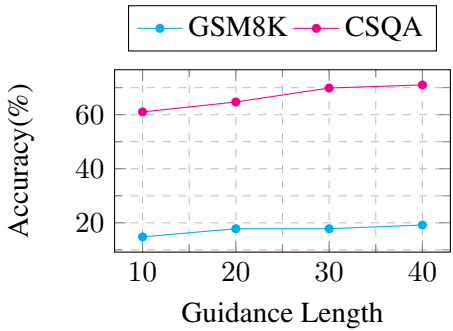

To investigate the optimal amount of assistance a teacher should provide to a student, or in other words, to determine the most suitable guidance length for maximizing student benefits, we conducted experiments. Intuitively, one might assume that the more a teacher model outputs, the higher the accuracy of the model's response. However, longer guidance tends to weaken the model's acceleration effect. Therefore, we plotted a line graph showing the changes in accuracy as the guidance length varied from 10 to 40, in intervals of 10, as depicted in Figure \ref{fig:length_range}. Based on our experiments, we established that for the GSM8K dataset, the teacher model should output the first 40 tokens of the answer. For the CSQA dataset, which generally requires shorter responses, we set the teacher model to output the first 30 tokens.

\subsubsection{How few-shot exemplars affect students?}
\label{append:few-shot-num}
\begin{figure}[!h] 
\centering 

\begin{tikzpicture} 

\begin{axis}[
    width=6cm, 
    height=4cm, 
    xlabel=Number of Few-shot Exemplars, 
    ylabel= Accuracy(\%), 
    tick align=inside, 
    legend columns=-1,
    legend style={at={(0.5,1.1)},anchor=south}, 
    minor tick num=1,
    grid=both,
    grid style=dashed
    ]

\addplot[color=cyan,mark=*,mark size=1.5pt,] coordinates{(1,10.38)(3,17.44)(5,17.81)(8,19.18)};
\addlegendentry{$\textrm{GSM8K}$}
\addplot[color=magenta,mark=*,mark size=1.5pt,] coordinates{(1,65.36)(3,68.71)(5,68.71)(7,69.86)};
\addlegendentry{$\textrm{CSQA}$}

\end{axis}
\end{tikzpicture}
\caption{Performance of different few-shot exemplar number }
\label{fig:few_shot}
\end{figure}
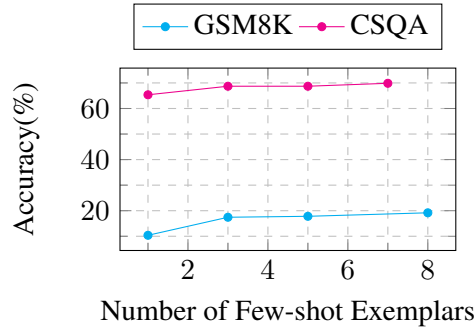

In this investigation, we focused on the role of few-shot exemplars in influencing the performance of student models. We used a range of few-shot exemplars, provided through Manual-CoT \cite{DBLP:conf/nips/Wei0SBIXCLZ22}, acros GSM8K and CSQA. The variation in accuracy with the change in the number of few-shot exemplars is illustrated in Figure \ref{fig:few_shot}. 
As depicted in Figure \ref{fig:few_shot}, it can be observed that the performance of the student models improves with the increase in the number of exemplars. Thus, in this experiment, we select 8-shot for GSM8K and 7-shot for CSQA in line with the Manual-CoT.


\section{Related Work}
\label{sec:related_work}
The aspect most relevant to our work is knowledge distillation\cite{DBLP:journals/corr/abs-2310-16944,jiang2023lion,vicuna2023,li-etal-2023-symbolic,DBLP:conf/acl/ChenGBS023}. Knowledge distillation involves condensing the expertise from larger models into more compact versions, thereby ensuring these smaller models retain high efficiency while still achieving impressive performance. 
\citet{li-etal-2023-symbolic} introduces Symbolic Chain-of-Thought Distillation (SCoTD), which trains a smaller "student" language model using the outputs (reasoning chains) of a larger "teacher" model. The teacher model first generates multiple reasoning chains for a given task, showcasing step-by-step problem-solving. The student model then are fine-tuned on these examples, essentially mimicking the teacher's reasoning process. This training enables the student model to perform complex reasoning tasks more effectively, despite its smaller size. 
\citet{DBLP:journals/corr/abs-2310-16944} presents a distillation framework named Distilled Direct Preference Optimization (dDPO) result in a 7B model named ZEPHYR. The method comprises three steps: (1) distilled supervised fine-Tuning (2) AI feedback through preferences  and 
(3) distilled direct preference optimization. Different from knowledge distillation, our work focuses on leveraging the knowledge of larger models to enhance the overall efficiency and performance. The proposed GKT framework circumvents the usual requirements of producing a distillation dataset from a teacher model or fine-tuning the student model, streamlining the knowledge transfer process.

Another line of study that is related to our work is speculative decoding \cite{DBLP:conf/icml/LeviathanKM23}. Speculative decoding uses two models: the original target model and a much smaller approximate model. The smaller model handles autoregressive sampling, while the larger assesses the output. Simple tokens are generated by the smaller model, with complex tokens handled by the larger. 
Based on speculative decoding, \cite{DBLP:journals/corr/abs-2311-08252} proposed Retrieval-Based Speculative Decoding (REST) which combines speculative decoding with retrieval techniques. Instead of using a smaller LM for draft generation, REST bypasses the need for an additional small LM by retrieving draft tokens from a pre-built datastore containing context-continuation pairs. These drafts are then verified by a large LM. 
Recently, \citet{DBLP:journals/corr/abs-2307-15337} proposed skeleton-of-thought (SoT). SoT first guides LLMs to first create a concise ``skeleton'' of an answer and then fill in each point of the skeleton in parallel, speeding up the response process. \citet{DBLP:journals/corr/abs-2305-08848} introduces Super In-Context Learning (SuperICL), a method enhancing the performance of LLMs by integrating them with smaller, locally fine-tuned models. These smaller models, acting as plug-ins, provide task-specific knowledge and predictions. The process involves fine-tuning a small model on task-specific data, using it to generate predictions and confidence scores for in-context examples, and combining these with the LLM's general language understanding. 
While SuperICL also integrates LLMs and LMs, it primarily focuses on enhancing model performance without considering the efficiency of the framework. This focus on performance enhancement can even negatively impact overall efficiency. Whereas, GKT focuses on finding the optimal balance between efficiency and effectiveness

\section{Conclusion}


We introduce the innovative Guidance-based Knowledge Transfer (GKT) framework, designed to leverage the knowledge of larger models to enhance the efficiency and performance of smaller models, while maintaining the flexibility of personalized generation settings, allowing users to freely adjust the generation parameters. The unique collaborative framework of GKT seamlessly integrates into cloud-edge architectures, deploying smaller models on edge devices to minimize data transmission delays and expedite response generation. 
Our results demonstrate remarkable improvements: a maximum accuracy increase of 14.18\% and a 10.72$\times$ speed-up on the GSM8K dataset, and a 14.00\% accuracy enhancement with a 7.73$\times$ speed increase on the CSQA dataset. Moreover, when utilizing ChatGPT as teacher model and Llama2-70B as the student model, we can achieve 95.00\% of ChatGPT's performance at 52\% of the cost.  GKT signify major strides in performance metrics, combining accuracy with computational speed-ups - all wrapped up in a ``cheap and cheerful'' package.

\section*{Limitation}
\label{append:limit}
The Guidance Knowledge Transfer (GKT) framework exhibits certain limitations compared to traditional knowledge distillation methods. Knowledge distillation primarily relies on a pre-trained smaller model, often leading to faster inference speeds. However, the GKT framework still depends on a large model to generate guidance prompts during inference. This approach aims to reduce the limitations brought about by fine-tuning through performance loss during inference while supporting customized configurations of the framework. 

Moreover, the length of guidance prompts generated by the GKT framework for different datasets is specific. Although the generation method is universally applicable, finding a universally appropriate length for guidance prompts that suits various scenarios remains a challenge. This means that additional adjustments and optimizations may be necessary for different datasets and application contexts to ensure optimal performance. Therefore, one of the future research directions is to explore how to more effectively determine the appropriate lengths of guidance prompts to enhance the universality and flexibility of the GKT framework.

\bibliography{custom,anthology}

\clearpage

\appendix
\section*{Appendix}
\section{Dataset Statistics}
\label{append:dataset}
The detailed dataset statistics can be found in Table \ref{tab:dataset}
\begin{table}[h]
\centering
\scalebox{0.8}{
\begin{tabular}{ccc}
\toprule
Dataset                      & GSM8K                           & CSQA                      \\ \midrule
\#Instance                & 1319                            & 1221                      \\
Average input length & \multirow{2}{*}{47}                              & \multirow{2}{*}{28}                        \\
(Words)  &                               &                         \\
Answer Format             & int                             & str                       \\
\multirow{2}{*}{Question Format}             & grade school       & single-choice    \\
                                    & math problems      &  question    \\
Few-shot                  & 8                               & 7                         \\ \bottomrule
\end{tabular}
}
\caption{Dataset statistics}
\label{tab:dataset}
\end{table}


\section{Hyperparameters}
The detailed hyperparameter settings can be found in Table \ref{tab:param}
\label{append:param}
\begin{table}[!h]
\centering
\scalebox{0.85}{
\begin{tabular}{ll}
\hline
Parameters              & Value \\ \hline
batch size (Bloom 7B)   & 32    \\
batch size (Llama2 13B) & 24    \\
batch size (Llama2 70B) & 10    \\
top\_p                  & 0.9   \\
temperature             & 0.8   \\
max\_seq\_len           & 1024  \\ \hline
\end{tabular}
}
\caption{Hyperparameters}
\label{tab:param}
\end{table}

\section{Speculative Decoding VS GKT}
\label{append:vs}
We provide a more comprehensive analysis of the deployment of speculative decoding and GKT within a cloud-edge framework in this section. For the Llama model, with a vocabulary size of 32K, the number of bits (\( b \)) required to store each token can be calculated using the formula:

\[
b = \lceil \log_2(N) \rceil
\]

where \( \lceil x \rceil \) denotes the ceiling function. Therefore, the number of bits needed to store each token is 15 bits.

In the scenario of GSM8K (40->300), for the GKT framework, to respond to a user request, it only needs to transmit 40 tokens of guidance prompts from the cloud LLM. However, for speculative decoding, the model needs to transmit 300 * 2 = 600 tokens (the small model transmits the draft token to the cloud LLM, and the LLM sends back the validated tokens to the small model).

In a network environment with limited bandwidth, say 5K per second, transmitting a 15-bit token takes about 0.003 seconds. Therefore, for GKT, transmitting 40 tokens would only require 0.12 seconds. However, for speculative decoding, it would take 1.8 seconds.
A significant advantage of the GKT framework over speculative decoding lies in its ability to control transmission losses by adjusting the length of the prompt. In contrast, the speculative decoding framework needs to transmit all tokens output by the LLM. Although speculative decoding can preserve the performance of LLM, it cannot be efficiently deployed in cloud-edge scenarios. On the other hand, GKT offers flexible applicability in these settings.

\section{Example for GKT framework}
\label{append:GKT_example}
\begin{figure}[t]
    \centering
    \scalebox{0.9}{
    \includegraphics[width=1\linewidth]{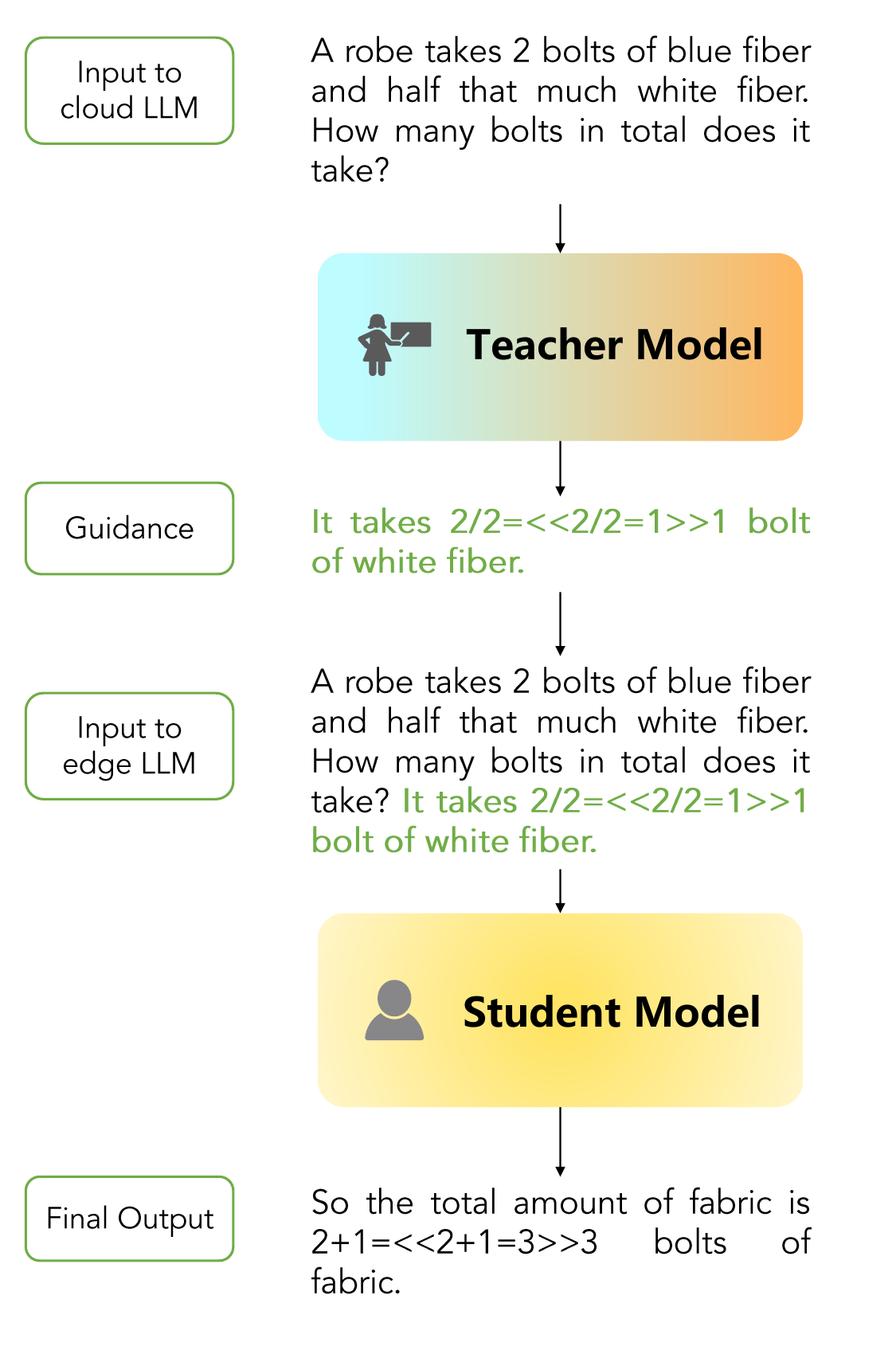}}
    \caption{Example for GKT framework}
    \label{fig:GKT_example}
\end{figure}

An concrete example for GKT framework can be found in Figure \ref{fig:GKT_example}, where we show the input and output of the cloud teacher LLM and the edge student LLM.

\section{Prompt used for GSM8K and CSQA}
\label{append:prompt}
\begin{figure*}
    \centering
    \includegraphics[width=0.85\linewidth]{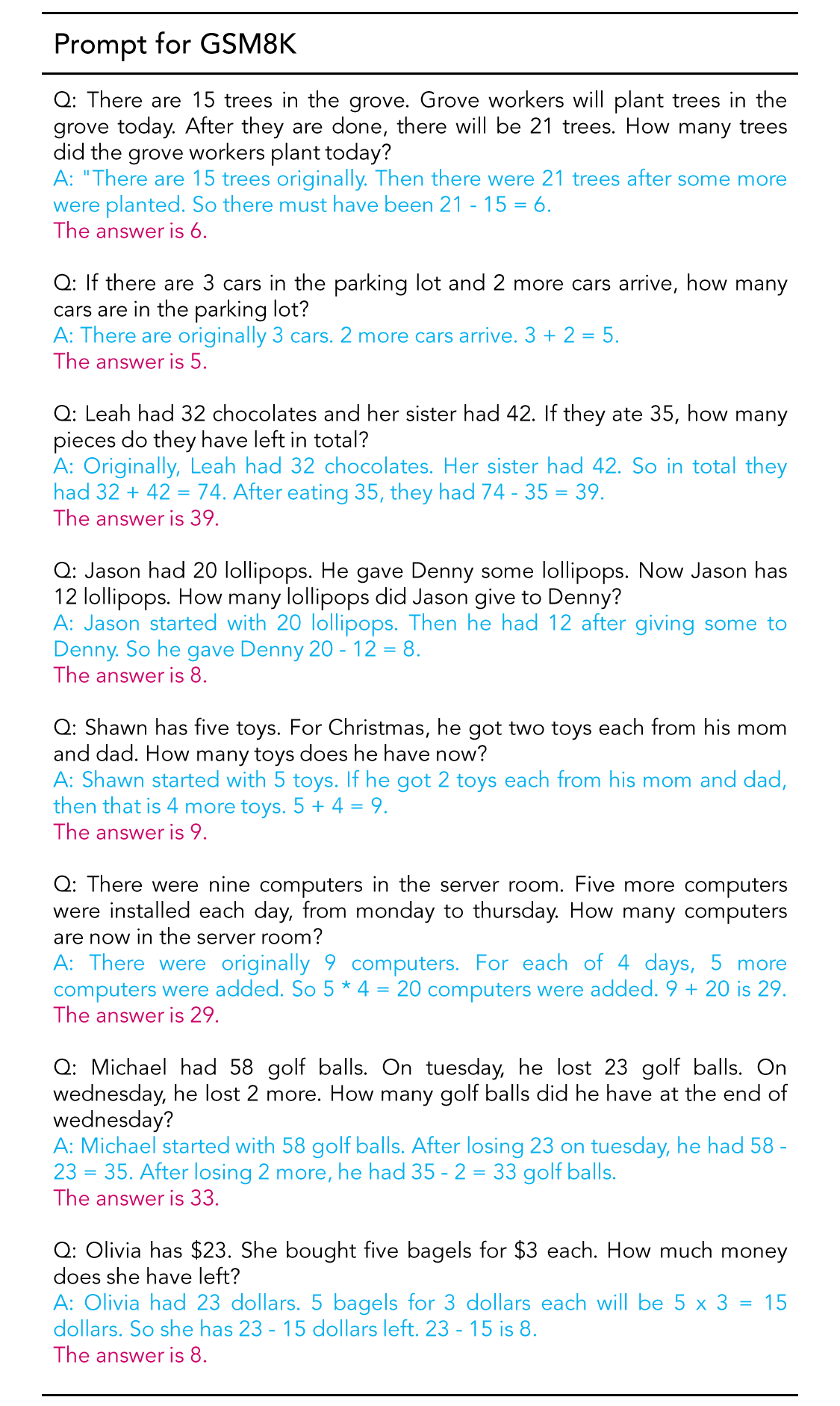}
    \caption{The prompt used for GSM8K}
    \label{fig:prompt_GSM8K}
\end{figure*}

\begin{figure*}
    \centering
    \includegraphics[width=0.85\linewidth]{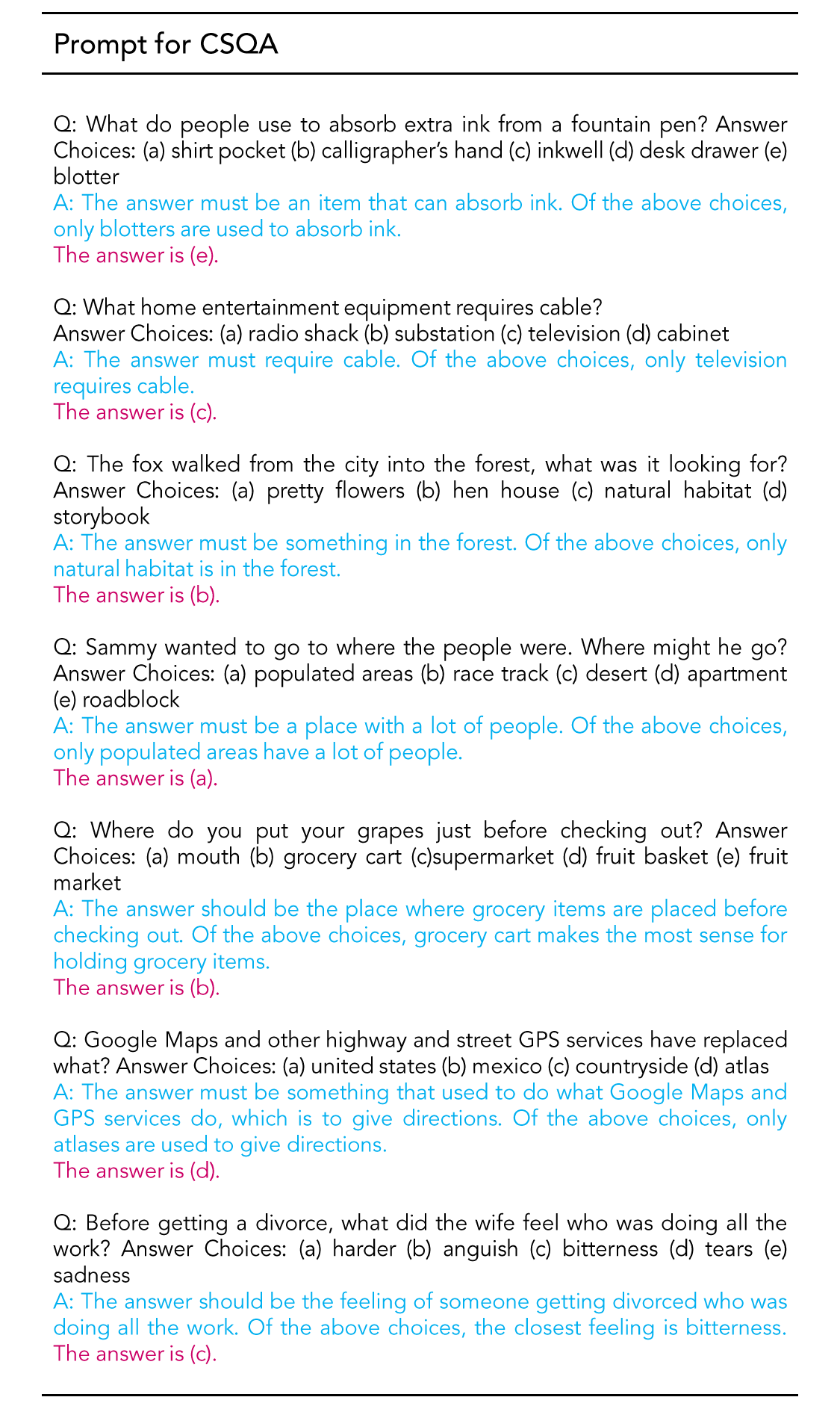}
    \caption{The prompt used for CSQA}
    \label{fig:prompt_CSQA}
\end{figure*}

The prompt used for GSM8K and CSQA can be found in Figure \ref{fig:prompt_GSM8K} and Figure \ref{fig:prompt_CSQA} respectively.

\end{document}